# Coupled Intrinsic and Extrinsic Human Language Resource-Based Query Expansion

B. Selvaretnam
Faculty of Computing & Informatics
Multimedia University
bhawani@mmu.edu.my

M. Belkhatir
Faculty of Computer Science
University of Lyon
belkhatir@univ-lyon1.fr

## ABSTRACT
Poor information retrieval performance has often been attributed to the query-document vocabulary mismatch problem which is defined as the difficulty for human users to formulate precise natural language queries that are in line with the vocabulary of the documents deemed relevant to a specific search goal. To alleviate this problem, query expansion processes are applied in order to spawn and integrate additional terms to an initial query. This requires accurate identification of main query concepts to ensure the intended search goal is duly emphasized and relevant expansion concepts are extracted and included in the enriched query. Natural language queries have intrinsic linguistic properties such as parts-of-speech labels and grammatical relations which can be utilized in determining the intended search goal. Additionally, extrinsic language-based resources such as ontologies are needed to suggest expansion concepts semantically coherent with the query content. We present here a query expansion framework which both capitalizes on linguistic characteristics of user queries and ontology resources for query constituent encoding, expansion concept extraction and concept weighting. A thorough empirical evaluation on real-world datasets validates our approach against unigram language model, relevance model and a sequential dependence-based technique.

## Categories and Subject Descriptors
H.3.1 [**Content Analysis and Indexing**]: Linguistic Processing
H.3.1 [**Information Search and Retrieval**]: Query Expansion

## 1. INTRODUCTION
Several query expansion efforts have emerged over the years in order to improve retrieval performance. Retrieval performance is often impeded by the query-document vocabulary mismatch problem which remains prominent due to the varying styles of writing by users as well as the semantic ambiguity that is present in natural language. In the context of search engines, query expansion involves evaluating a user's input (what words were typed into the search query area, and sometimes other types of data) and expanding the search query to match additional documents. The query expansion process requires the comprehension of the intended human search goal through: i) subsequent identification of key concepts prior to generating additional concepts enriching the query and, ii) assessing the importance of the original and added concepts through robust weighting schemes.

As far as the first issue is concerned, current research for key concept identification tasks (e.g. Huston & Croft 2014) incorporate a combination of syntactical analysis and statistical mining. Nouns are assumed to be of significance (Cao et al. 2005), while more often all non-stop words query terms are assumed to be representative of query content (Voorhees 94), in the same fashion as unigram models in early information retrieval efforts. Several other works on key concept extraction in documents (Frank et al. 1999) and queries (Lioma & Ounis 2009, Bendersky & Croft 2008) also emphasize on noun phrases and key concepts are established through the examination of frequency of occurrence of terms in documents, n-grams, query logs, etc. It is, however, inherent that query elements possess two different functionalities that are disregarded: concepts either characterize the content according to the search goal or are used to connect query elements. It is therefore important to consider that query constituents can take on several roles which if recognized and taken into account appropriately would render a more accurate understanding of the intended search goal.

Additionally, the nature of natural language dictates that there are intrinsic relationships between adjacent and non-adjacent concepts that highlight semantic notions pertaining to a search goal. Earlier works mostly fail to fully capitalize on these relationships between query terms which if considered appropriately would improve retrieval performance. One might argue that both adjacent and non-adjacent dependencies can be taken into account through full dependence modeling. This will however prove costly, in especially long queries, as multiple concept pairs will be derived, from which possibly a large number would not be very meaningful. Query expansion based on these pairs would generate unrelated concepts which in turn cause digression from the original search goal. We hypothesize that both adjacent and non-adjacent association among query concepts can be effectively capitalized from syntactical dependencies within queries. This then translates into meaningful query concept pairing for expansion.

Furthermore, rather than using statistical techniques for generating additional query concepts, the use of language-based resources such as ontologies allow spawning then integrating in the original query expansion concepts that are semantically consistent with its content.

A crucial element in query expansion is the process of weighting the original and expansion concepts to adequately reflect the search goal of a query. Thus far, state-of-the-art methods have placed most emphasis on the frequency of concepts within a document corpus either through simplistic term and document frequency computation (such as in Song et al. 2008) or through supervised learning mechanisms with multiple features but are also centered on the frequency of the concept within a variety of sources such as n-grams and query logs (Paik & Oard, 2014). The drawback of such models is that key concepts are established based on the statistical occurrence of a concept which is not necessarily reflective of the search goal of a query. Instead, we believe concepts should be given due emphasis based on the conceptual role they play in representing the information need. There are several efforts that attempt to model query constituent weights using genetic algorithms (Yang et al. 2003, Kraft et al. 1995, Simon & Sathya 2009). However, these efforts do not consider the role of the concepts within a query.

We present a query expansion framework that consists of three components: (i) a linguistically-motivated scheme for recognizing and encoding significant query constituents that characterize the intent of a query (Selvaretnam & Belkhatir 2016); (ii) a module for the generation of potential expansion terms based on a extrinsic lexical resource (ontology) that capitalizes on grammatically linked base pairs of query concepts; (iii) a robust weighting scheme reconciling original and expansion concepts that is reflective of the role types of query constituents in representing an information need. After analyzing the state-of-the-art solutions in Section 2, we provide a detailed analysis of the framework components in Section 3 and present their algorithmic instantiation in Section 4. We present the evaluation of our framework on large-scale real-world datasets in Section 5.

## 2. ANALYSIS OF LANGUAGE RESOURCE-BASED QUERY EXPANSION FRAMEWORKS

In Figure 1, the general process flow of a language resource based query expansion framework is illustrated. Through the example processing of an original query ("*how to repair a car*"), we illustrate the three major query handling steps involved: (i) linguistic processing, (ii) sense disambiguation and, (iii) lexical-semantic term pooling.

With regards to linguistic processing, most related works subject queries to minimal linguistic analysis where they are stemmed and stop words removed prior to the process of expansion. There are several efforts that recognize the importance of identifying phrases that occur in a query instead of treating all terms in a query independently (Liu et al. 2004, Park et al. 2011, Pinter et al. 2016). They rely on an ontology to perform syntactical analysis in order to identify proper names and dictionary phrases but it is not clear whether they cover all crucial non-compositional phrases such as phrasal verbs, idioms, collocations, acronyms, etc. Also, they do not utilize syntactical dependencies to identify simple and complex phrases that are not found in the dictionary but instead devise a simple grammar that relies on the existence of noun phrases and content words (i.e. non-stop words).

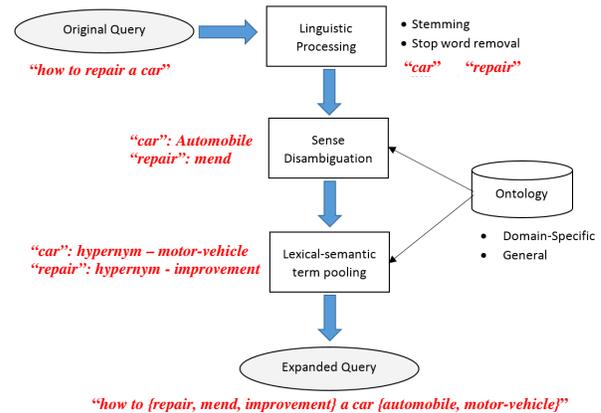

**Figure 1: Organization of Language Resource-Based Query Expansion Frameworks**

Among the approaches for conducting word sense disambiguation, most authors handle only nouns and fail to consider verb senses in their approach. This may not be wise as verb senses could bring further improvement in understanding the intended meaning of the query. Liu et al. (2005) perform word sense disambiguation in short queries, choosing the right sense for a word in its occurring context. Noun phrases of the query are determined and compared against synonyms and hyponyms from WordNet to identify word senses. If it is not possible to determine the word sense in this manner, then a guess is made based on the associated frequency of use (i.e. the number of times a sense is tagged) available in WordNet. The approach of utilizing the "*first sense heuristic*" (i.e. choosing the first or predominant sense of a word) is reiterated by McCarthy et al. (2007) as a method that is used by most state-of-the-art systems as a back-off method when text-based disambiguation is unsuccessful. Broadly speaking, sense disambiguation methods are far from perfect resulting in multiple research works that attempt to improve the accuracy of disambiguation through improved measures of relatedness (e.g. Patwardhan et al. 2007, Navigli & Velardi 2003) and via information extracted from external sources (e.g. Mihalcea 2007).

As far as term pooling is concerned, based on the derived sense of the query concepts, lexical and semantically related expansion concepts are extracted from an ontology. Query expansion frameworks have often relied on lexical resources (i.e. dictionaries, thesauri etc.) for the purpose of suggesting semantically related terms and sense disambiguation. Often, these resources are ontologies, either domain specific or general. Query expansion using domain-specific ontologies is more suitable for static document collections. For Web collections, the ontologies would have to be frequently updated because the collections on the web are more dynamic in nature. Terminologies in domain-specific ontologies are less ambiguous, therefore queries for narrower search tasks can be expanded with a higher chance of accuracy. General ontologies would be suitable for broad queries (Bhogal et al., 2007). In previous research efforts involving external knowledge sources, Voorhees (1994), Liu et al. (2004), Chauhan et al. (2012), Hollink et al. (2007), Mestrovic & Cali (2016) utilize general ontologies; Radhouani et al. (2006), Tudhope et al (2011), Alejandra Segura et al. (2011), Bhogal et al. (2013), Abdulla (2016) use domain-specific ontologies while Tuominen et al. (2009) evaluate their framework with both general and domain-specific ontologies. Query structures and varying query lengths (Selvaretnam et al. 2013) were not differentiated during the process of expansion in these efforts. However, retrieval performance in long queries may be improved through the approach of Radhouani et al. (2006). In the latter, an ontology is used to identify semantic *dimensions* of medical queries rather than extract related terms and a document is considered relevant if it contains one or more dimensions of the query. Unfortunately, identification of dimensions of a query is not straightforward in general queries as the terms utilized in such queries and those found in general ontologies are highly ambiguous. For non-domain specific queries, this approach of identifying dimensions could be likened to the identification of main query concepts. Main query concepts can be identified in long queries, especially, allowing documents that contain one or more of the main concepts to be deemed as relevant. Voorhees (1994), Liu et al. (2004), Chauhan et al. (2012), Alejandra Segura et al. (2011), Bhogal et al. (2013) use only one source of potential expansion terms, that is the external knowledge base. Tuominen et al. (2009) utilize the ontologies published in the ONKI Ontology Service. All works include lexically related terms, Chauhan et al. (2012) use synonyms while Voorhees (1994) and Liu et al. (2004) additionally utilize hyponyms. Bhogal et al. (2013) make use of both synonyms and hypernyms. Hypernyms and meronyms are also integrated in Voorhees' model whilst Liu et al. (2004) consider compound terms as well. Alejandra Segura et al. (2011) use common ontological relations and ontological relations specific to the Gene domain-specific ontology in their expansion process. The produced mix of concepts may not work well when terms which are of contrasting relations to a query topic (e.g. generalization and specialization) are utilized together causing the range of related documents to be less precise. Koopman (2016) model the entities and relationship types based on domain specific ontologies and attempt to infer relevance through a graph inference model. The proposed model works to improve recall and precision for domain-specific queries, it is unable to perform well for concepts which are more generic in nature. To conclude, the current approaches do not handle non-compositional phrases and rely on simple syntactical analysis which focuses only on noun phrases in order to identify the key concepts in a query. They also do not sufficiently address the process of lexical-semantic term pooling where the relationships between terms are not considered prior to term pooling. The extracted terms are also not given adaptive weight to accurately reflect the role of the original and expansion terms.

## 3. DESCRIPTION OF FRAMEWORK COMPONENTS

An enhanced query expansion framework is proposed to address the unresolved issues highlighted in Section 2. Figure 2 shows the stages of processing a query within the proposed query expansion framework and reflects the intermediate output of each phase for a sample query "mild yeast infection".

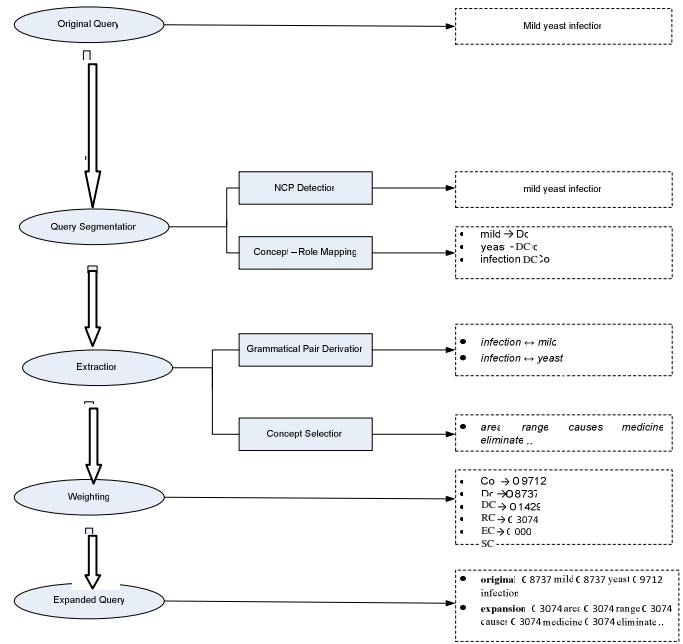

**Figure 2: General organization and processing flow of a query**

The framework consists of three phases, namely query segmentation, extraction and weighting. The query segmentation phase consists of the non-compositional phrase (NCP) detection and concept-role (CR) mapping processes. The NCP detection process serves to identify and preserve NCPs in their full form so as to accurately capture the original intent. The CR mapping phase aims to derive relationships between query terms and assign a role-type to each term based on its functionality. The extraction phase focuses on the elicitation of lexical-semantic related terms from external knowledge sources given a base term pair. Lastly, the weighting phase handles the generation of term weights for each original and expansion term in an effort to fairly place emphasis on terms that truly reflect the original search goal. Each phase is described in greater detail in Sections 3.1 to 3.3.

## 3.1 Query Segmentation

### 3.1.1 Non-Compositional Phrase Detection

Queries are typically made up of compositional and non-compositional phrases (NCPs). NCPs have to be isolated in order to preserve the desired intent of the user query. Among the various types of NCPs that exist within the English language, phrasal verbs, idioms, collocations, proper names and acronyms/abbreviations are of particular interest as they contribute directly towards a search goal.

### 3.1.2 Concept Role-Mapping

A query consists of constituents that play the role of either content or function words. Content words (e.g. nouns, verbs, adjectives and adverbs) refer to actions, objects and ideas in the everyday world. Function words (e.g. determiners, auxiliary pronouns, conjunctions, prepositions), on the other hand, express grammatical features like mood, definiteness and pronouns reference or facilitate grammatical processes like rearrangement, compounding and embedding. Typically, nouns represent the subject of a sentence through names of persons, places, things and ideas whilst verbs are actions which are representative of a predicate. Other constituents such as determiners and adjectives modify nouns whilst adverbs generally alter verbs. Similarly, prepositions take on a modifying role, acting as adjectives/adverbs, by describing a relationship between words. Conjunctions, on the other hand, connect parts of a sentence or specifically the content words. However, this crude distinction of content and function words based on parts-of-speech renders all nouns, verbs, adjectives and adverbs as key concepts. Closer scrutiny of queries reveal that content words (e.g. nouns, adjectives) sometimes serve as modifiers and complements which complete the meaning of a query, thus may not necessarily be key concepts. To illustrate, we refer to the example query "*coping with overcrowded prisons*". The parts-of-speech are identified where "*coping*" is a verb, "*with*" a preposition, "*overcrowded*" an adjective and "*prisons*", a noun. The content words in this case would be "*coping*", "*overcrowded*" and "*prisons*" whilst "*with*" takes the role of a function word. However, it would be more precise to say that "*prisons*" is the key concept of this query; "*overcrowded*", an adjectival modifier, describes the state of the prisons whilst the word "*with*" connects both words to reflect the act of dealing with prison conditions. The existence of modifiers and complements indicates that there is a need to further specify the role of the various parts-of-speech to more accurately capture the intent of a given query for more effective query expansion. It is our postulation that query constituents may be categorized into four types of concepts (which we refer to as *role-type* from this point on) that characterize the role of the constituents within a query: (i) Concepts-of-Interest (*CoIs*) are the key concepts in a query indicative of user intent; (ii) Descriptive Concepts (*DCs*) describe the key concepts in further detail; (iii) Relational Concepts (*RCs*) provide the link between query concepts; (iv) Structural Concepts (*SCs*) are stop words that help form the structure of a query.

A simplistic approach of role type categorization based only on parts-of-speech would result in the mislabeling of role type. Both adjacent and non-adjacent pairs of concepts should be considered in the concept-role mapping process. Thus, typed dependencies are used to discover links between individual query constituents (Covington 2001) and consist of three main grammatical relation groupings: auxiliary, argument and modifier. The auxiliary dependencies consist of relations between words which are connected by stop words (e.g. conjunctions). The argument dependencies draw attention to subjects, objects, and complements within a sentence. The modifier dependencies highlight several parts-of-speech based modifiers (e.g. adjectival, noun compound, etc) which describe and complete the meaning of other constituents in a sentence.

The concept-role mapping process entails 2 consecutive steps: (i) Extraction of grammatical relations between query constituents and determination of *head* (i.e. word upon which everything in a phrase is centered) and *dependent* (i.e. every other word that is associated to the *head*)"; (ii) Assignment of concept roles depending on the functionality of the constituents linked by the grammatical relations.

All main grammatical relations (55 relations) defined in the Stanford Typed Dependency Scheme (Marneffe & Manning, 2008) are considered in this framework inclusive of auxiliaries, arguments and modifiers, as well as several sub-relations that further specify the main relations. The defined grammatical relations are used to infer the appropriate role-type assignment for query concepts in three categories (e.g. arguments, modifiers, and auxiliaries) such as shown in Tables 1a, 1b, 1c.

**Table 1a: Grammatical Relations for CR Mapping (Arguments)**

| Grammatical Relations | Description | Concept Roles | |
|---|---|---|---|
| | | Head | Dependent |
| Coordination | Relation between an element of a conjunct and the coordinating conjunction word of the conjunct. | CoI | RC |
| Adjectival complement | Adjectival phrase which functions as the complement. | DC | CoI |
| Clausal complement | Dependent clause with an internal subject which functions like an object of the verb, or adjective. | DC | CoI |
| Open clausal complement | Clausal complement without its own subject, whose reference is determined by an external subject. | DC | CoI |
| Complementizer | Word introducing a clausal complement. (subordinating conjunction "that" or "whether"). | CoI | RC |
| Direct object | Noun phrase which is the (accusative) object of the verb in a VP. | DC | CoI |
| Indirect object | Noun phrase which is the (dative) object of the verb in a VP. | DC | CoI |
| Object of a preposition | Head of an NP following the preposition, or the adverbs | RC | CoI |
| Marker | Word introducing an adverbial clausal complement. | CoI | RC |
| Relative | Head word of the WH-phrase introducing a relative clause. | CoI | RC |
| Nominal subject | NP being the syntactic subject of a clause. | DC | CoI |
| Passive nominal subject | NP which is the syntactic subject of a passive clause. | DC | CoI |
| Clausal subject | Clausal syntactic subject of a clause, i.e., the subject is itself a clause. | DC | CoI |
| Clausal passive subject | Clausal syntactic subject of a passive clause. | DC | CoI |
| Expletive | Captures an existential "there". | RC | RC |
| Prepositional complement | Used when the complement of a preposition is a clause, prepositional phrase or adverbial phrase. | RC | RC |
| Rreconjunct | Relation between the head of an NP and a word that appears at the beginning\bracketing a conjunction. | CoI | RC |

**Table 1b: Grammatical Relations for CR Mapping (Modifiers)**

| Grammatical Relations | Description | Concept Roles | |
|---|---|---|---|
| | | Head | Dependent |
| Noun compound modifier | Any noun that serves to modify the head noun | CoI | DC |
| Adjectival modifier | Any adjectival phrase that serves to modify the meaning of an NP. | CoI | DC |
| Prepositional modifier | Any prepositional phrase that serves to modify the meaning of a verb, adjective, noun, or another prepositon. | DC | CoI |
| Abbreviation modifier | Parenthesized NP that serves to abbreviate an NP | CoI | CoI |
| Appositional modifier | NP immediately to the right of an initial NP that serves to define or modify it. | CoI | CoI |
| Adverbial clause modifier | Clause modifying a verb (temporal clause, consequence, conditional clause, etc.). | DC | CoI |
| Purpose clause modifier | Clause headed by "(in order) to" specifying a purpose. | DC | CoI |
| Numeric modifier | Any number phrase that serves to modify the meaning of a noun. | CoI | DC |
| Element of compound number | Part of a number phrase or currency amount. | CoI | DC |
| Possession modifier | Holds between the head of an NP and its possessive determiner, or a genitive's complement. | CoI | CoI |
| Phrasal verb particle | Identifies a phrasal verb, and holds between the verb and its particle. | DC | CoI |

| Parataxis | Relation between the main verb of a clause and other sentential elements, such as a sentential parenthetical, or a clause after a ":" or a ";". | CoI | RC |
| Punctuation | Used for any piece of punctuation in a clause, if punctuation is being retained in the typed dependencies. | CoI | SC |
| Referent | A referent of the head of an NP is the relative word introducing the relative clause modifying the NP. | CoI | RC |
| Controlling subject | Relation between the head of a open clausal complement and the external subject of that clause. | DC | CoI |

**Table 1c: Grammatical Relations for CR Mapping (Auxiliaries)**

| Grammatical Relations | Description | Concept Roles | |
|---|---|---|---|
| | | Head | Dependent |
| Auxilliary | Non-main verb of a clause, e.g. modal auxiliary, "be" and "have" in a composed tense. | CoI | RC |
| Passive auxiliary | Non-main verb of a clause which contains the passive information. | CoI | SC |
| Copula | Relation between the complement of a copular verb and the copular verb. | CoI | RC |
| Agent | Complement of a passive verb introduced by the preposition "by" and does the action. | RC | CoI |
| Attributive | Complement of a copular verb such as "to be", "to seem", "to appear". | RC | RC |
| Conjunction | Relation between 2 elements connected by a coordinating conjunction. | CoI | CoI |
| Determiner | Relation between the head of an NP and its determiner. | CoI | SC |
| Predeterminer | Relation between the head of an NP and a word that precedes and modifies the meaning of the NP determiner. | CoI | RC |

In some cases, two issues, namely untagged and ambiguous concept-role mapping, may occur depending on the parse output of queries. Query concepts are untagged when the most generic grammatical relation, i.e. undefined (*undef*), is highlighted due to the inability to produce a more precise relation between a pair of terms. A statistical approach making use of the frequency of a concept to determine untagged concept role types is proposed. The frequency of terms can be obtained from Google as it provides the frequency of terms across all indexed documents on the Web. This is in line with most existing approaches of identifying key concepts where the assumption is that frequently occurring concepts are key concepts.

Thus, three rules are derived to handle untagged concepts:
(i) if a term is tagged in one of the highlighted grammatical relations, its corresponding role is selected;

(ii) the term that occurs more frequently is the *CoI* and the other is a *DC* unless a role has been previously assigned according to the first rule;

(iii) if both terms are equally frequent, both terms are tagged as *CoIs* unless a role has been previously assigned according to the first rule.

We consider an application of these rules given the example query "*United States control of insider trading*" in Table 2. A parser derives three relations among the query terms, i.e. noun compound modifier (*nn*), undefined (*undef*) and preposition (*prep_of*). The relationship between the terms "*United States*" and "*control*" is undefined (i.e. *undef*). According to the first rule, the term "*control*" is assigned the role-type *DC* from its occurrence in the relation *prep_of*. The term "*United States*" is originally tagged as *U*, which signifies that an appropriate role-type was not found. However, upon examination of its frequency in accordance to the second rule, the term "*United States*" is tagged with the role-type *DC*.

**Table 2: Handling Untagged Query Terms**

| Relation | Head | Role-Type | Dependent | Role-Type |
|---|---|---|---|---|
| nn | trading | CoI | insider | DC |
| **undef** | **United_States** | **U** | **control** | **U** |
| prep | control | DC | trading | CoI |
| prep | of | RC | - | - |

Ambiguous concept-role mapping occurs due to the nature of natural language queries in which adjacent and non-adjacent terms can be syntactically or semantically associated. This means that a term can be linked to one or more terms within a query, playing a different role in each partnership. To address this, the role-type tag of the more significant role-type is kept. When the contradiction is caused by either preposition or conjunction relation, the role-type tag of any other relation is retained. This is explained by the fact that these relations serve to concatenate terms and as such are assumed to be less significant than other relations. If the contradiction is between the preposition and conjunction relations, the role from the preposition relation is kept because prepositions connect terms that modify one another while conjunctions simply connect two terms. Thus, three rules are derived to handle ambiguous roles:
(i) the role-type tag of the more significant role is retained with *CoI* being the most significant role, followed by *DC*, *RC* and lastly *SC*.
(ii) if the contradiction is caused by conflicting relations, the role-type tag from the relation that has higher priority is retained, where all other relations are prioritized over the preposition and conjunction relations.
(iii) if the contradiction is between a preposition and a conjunction relation, the role-type tag from the preposition relation is retained where the preposition relation has higher priority than the conjunction relation.

The retention of the more significant role in the second case is illustrated for the query "*Iranian support for Lebanese hostage takers*" processed in Table 3. A parser derives two relations among the query concepts, i.e. adjectival modifier (*amod*) and prepositional modifier (*prep_for*). A contradiction in the role assignment is highlighted with the concept "*support*". The role derived from the relation *amod* (i.e. *CoI*) is then retained rather than the *DC* role-type suggested when considering the *prep_for* relation.

**Table 3: Handling Ambiguous Concept-Role Mapping**

| Relations | Head | Role-Type | Dependent | Role-Type |
|---|---|---|---|---|
| **amod** | **support** | **CoI** | Iranian | DC |
| amod | hostage_takers | CoI | Lebanese | DC |
| **prep_for** | **support** | **DC** | hostage_takers | CoI |
| prep_for | for | SC | - | - |

To illustrate the process of concept-role mapping, we again refer to the query, "*mild yeast infection*". The linguistic parser returns that "*mild*" is an adjective and "*yeast*" a noun that both modify the noun "*infection*". Referring to the set of concept role assignments, we map each query term to its role, i.e. "*mild*" is a *DC*, "*yeast*" is also a *DC* and "*infection*" a *CoI*.

## 3.2 Extraction

### 3.2.1 Selection of Base Terms

Identification of candidate expansion terms for the external language resource-based processing approach of query expansion revolves around determining the pool of related lexical-semantic terms of individual terms in a query. At this point, a question arises as to which terms within a query are of particular importance in the process of expansion. Each term within a query plays a certain role which either exhibits content or function. Thus, blind expansion on terms that do not represent the main content of the query would cause deterioration of retrieval performance. The functionality of each term within a query had been represented using four role-types through the concept-role mapping process defined in Section 3.1.2. Query terms which reflect the main goal of the query are annotated with the *CoI* and *DC* role-types. *CoIs* represent the key concepts in a query whilst terms labelled with the *DC* role-type further specify the main search goal.

Thus, for the purpose of extracting lexical-semantic related terms, only the *CoI* and *DC* role-types are used as *base terms*. Query terms which provide links between terms (i.e. the *RC* role-type) as well as stop words which form the structure of the query (i.e. the *SC* role-type) are not considered in the term pooling process. The constraint imposed on the terms to be selected as base terms is due to the nature of terms labelled based on the basic role-types. *CoI*s and *DC*s are open-class words which represent content such as nouns, verbs (transitive and intransitive) and adjectives. The open-class words category is represented in the form of ontologies consisting of hierarchical, equivalence and associative relations among the terms. This would mean that *CoI*s and *DC*s, which represent the main search goal are easily expandable through the traversal of ontologies. The set of term associations defined within the ontology models the multiple terms that form the topic of interest in a related document. On the other hand, *RC*s and *SC*s belong to the closed-class words category which consists of stop words such as determiners, conjunctions and prepositions. Modals (e.g. can, will, may), auxiliary verbs (e.g. be, have, do) as well as adverbs are also labelled as *RC*s or *SC*s. These groups of function words are not accepted as base terms as they do not possess semantic associations with other terms and are not representative of the search goal. The terms "*coping*", "*overcrowded*" and "*prisons*" are respectively annotated as *DC*, *DC* and *CoI* while the preposition "*with*" is labelled as an *RC* in the example query "*coping with overcrowded prisons*". Hence, although the query consists of four terms, only three of them are adopted as base terms for the term pooling process.

### 3.2.2 Disambiguation of Word Senses for Accurate Term Pooling

The identified base terms are central to the lexical-semantic term pooling process. They indeed are the only query terms which are expanded as they are the only components of the query which represent the search goals. However, each base term is prone to polysemy, a likely occurrence in natural language. This presents a problem in the term pooling process because polysemous terms have multiple meanings with variations found in and across the different parts-of-speech and as such leading to a different set of candidate expansion terms. For example, the polysemous term "*Java*" has three distinct sets of related terms in accordance to its intended meaning in a given context (cf. Table 4). If the wrong meaning of the term is assumed in the term pooling process, the expansion process would have caused a serious query drift. Therefore, all base terms should be disambiguated prior to extracting candidate expansion terms.

**Table 4: Possible related terms for the polysemous term "*Java*"**

| Term | Meaning | Related Terms |
|---|---|---|
| Java | An Indonesian island | land, dry land |
| Java | A caffeinated beverage | beverage, drink, food, nutrient |
| Java | A programming language | object-oriented, artificial language |

The Word Sense Disambiguation (WSD) process consists of two essential steps: i) the determination of all the different senses for every word relevant to a term and, ii) methods to assign the appropriate sense to a given term within a particular context.

The disambiguation technique applied for the lexical-semantic related term pooling process is the knowledge-based WSD for two main reasons. Firstly, this choice is based on the observation that when there are larger amounts of structure knowledge, WSD performance improves (Cuadros & Rigau, 2005). Secondly, although some may argue that knowledge sources such as ontologies are plagued by the data sparseness issue, there is a significant body of work aimed at the expansion of knowledge sources (e.g. Maree & Belkhatir 2011, 2013 2015). The *all-words* disambiguation approach is most suited here as the base terms are usually open class words and each query typically contains multiple *CoI/DC* labeled base terms. The semantic distance between query terms must be computed in order to annotate a term with its correct sense within a given query context. In line with the chosen mode of knowledge-based WSD, similarity measures which are based on lexical resources are considered in this work. The relatedness measures available for knowledge-based disambiguation are commonly achieved through the representation of ontologies as a graph followed by the computation of relatedness from a path between two terms. On the other hand, the *gloss-based* disambiguation method quantifies the number of common words or phrases identified as overlaps between term definitions. While both approaches have been applied on two prominent knowledge sources, WordNet and Wikipedia, a more significant amount of research has been focused on WordNet. Although Wikipedia is gaining prominence due to its ability to provide larger coverage, WordNet remains the preferred knowledge source for disambiguation (Fogarolli, 2011). The motivation behind this choice is the manual crafting of the WordNet ontology which lends greatly to the assurance of correctness in the annotation (Kuroda & Bond, 2010) and organization of explicit taxonomic relationships required in the term pooling process described next.

### 3.2.3 Term Pooling

A document discussing a particular topic is likely to contain terms which are lexically or semantically related. Classical lexical-semantic relations are term relationships which are hierarchical, equivalent or associative as shown in Figure 3.

Equivalence is recognized between terms which have similar meaning whilst associative relationships link related-to/associated terms of a particular topic. Hierarchical term relationships denote broader to narrower term relationships organized within multiple levels of superordination and subordination. The generic hierarchical relationship is the *is-a* link which connects a class and its members. Among the multiple existing lexical-semantic term relationships, the synonymy, hypernymy, hyponymy and *related-to* associations are considered in the term pooling process.

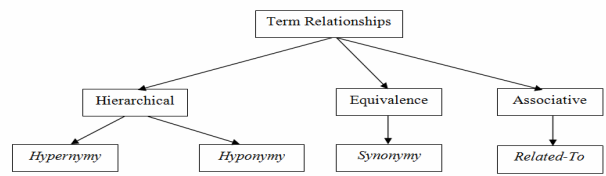

**Figure 3: Organization of Lexical-Semantic Term Relationships**

The sense disambiguated base terms are expanded through the extraction of lexical-semantic related terms. Since each term relationship denotes varying associations with one another, they cause redirection of the search goal in a very distinct manner. For example, the expansion of a term (e.g. "car") with its hypernym (e.g. "automobile") broadens the search whilst the inclusion of its hyponym (e.g. "Ford") focuses the search results to a narrower scope. The further the hierarchy is traversed up or down, the further the results are specialized or generalized. Combining the terms from these varying relations would be conflicting. As such, the candidate expansion terms are pooled separately to avoid mixed redirection of the search goal. Since related terms are extracted for all base terms for each query, multiple occurrences of identical expansion candidate terms are bound to occur which are duly filtered.

All levels in a hierarchy of each relation are traversed during the extraction process and the corresponding synonyms of each hypernym, hyponym and coordinate terms (which represent *related-to* associations between terms) are also retrieved from the WordNet ontology for optimal coverage. The semantic relatedness between the original queries and the related terms is then computed based on the Adapted Lesk (Pedersen et al., 2004) which is a gloss-based semantic relatedness measure. In cases where the accurate sense of a candidate expansion term cannot be determined, the most frequent sense of the term is assumed based on the first sense heuristics. In order to estimate the global relatedness of a candidate term to the entire query, the average relatedness is computed between each candidate expansion term and all original terms. However, the number of optimal terms to be included in the expanded query is empirically determined.

### 3.3 Concept Weighting:

The usefulness of a term within a query is not always determined solely by its functionality. Rather, retrieval performance is greatly enhanced by the inclusion of crucial terms where the common theme among the global pool of terms significantly impacts the query retrieval effectiveness. Intuitively it may be assumed that *CoIs* and *DCs* are more important than *RCs* and *SCs*, this assumption may not always hold true depending on the collective theme of the query terms. Thus, a technique that can randomly vary the term weights to optimize performance of a query is formalized through the use of a Genetic Algorithm (GA) formulation. GA is utilized to estimate the weights for each role type given a query for which the mean average precision (MAP) of retrieval is optimized. At this point, a fifth role type is introduced to characterize generated expansion concepts (*EC*). The GA optimization consists of the following: (i) Initialization of the population of chromosomes, elements encoding potential solutions that consist of five genes representing each role type; (ii) Implementation of a fitness function for evaluating each potential solution where the performance measure is based on the maximum MAP; (iii) Selection of the fittest individual of the current population (i.e. corresponding to the maximum MAP) which is used for reproduction. The parameters and optimal rates utilized in the GA implementation are determined through a tuning phase by varying the parameters on a training set of 50 queries until a steady state of optimized retrieval performance is achieved. Role-type weights are generated for 4 out of 5 role types identified, i.e. *CoI*, *DC*, *RC* and *EC*, with a boundary condition set to restrict weights within a 0 to 1 weight range. The *SC* role type is enforced with a null weight in order to eliminate emphasis on common terms. The global role-type weights are generated at each iteration and used on each query. Population is evolved for an optimal number of iterations as a terminating condition. Random perturbations to the population of chromosomes are inflicted through the process of mutation and cross-over. Concept scores for the example query "*mild yeast infection*" are given in Figure 2.

## 4. ALGORITHMIC INSTANTIATION

### 4.1 Query Segmentation

The algorithm below summarizes the implementation of the query segmentation module:

```
Input: Q ← The set of web queries
1: for all q in Q do
2:    ArrayList Concepts=breakQueriesIntoConcepts()
3:    String[] NCPList = getNCPList()
4:    if q is in NCPList
5:       Format query to highlight NCPs
6:    ArrayList RCP = getRelation&ConceptPairs()
7:    for all rcp in RCP do
8:       Assign role-type tag to each concept pair
9:    for all c in Concepts do
10:      ArrayList RTCP=getRelation&TaggedConceptPairs()
11:      for all rtcp in RTCP do
12:         String[] ConceptTag = getConceptTags()
13:         for all ct in ConceptTag do
14:            if ct is "untagged"
15:               Resolve according to 2 cases
16:                  Case 1: Frequency Dependence
17:                  Case 2: Inheritance Dependence
18:            if ct is "ambiguous"
19:               Resolve based on Ranked Relations
20:         end for
21:      end for
22:   end for
23:   end for
24: end for
```

As far as NCP detection is concerned, the queries are compared against the *NCPList*, a knowledge base of NCPs which was built by compiling all such phrases from multiple sources such as ontologies (e.g. WordNet) and standard lists found on the Web. If a query is in *NCPList*, it has to be appropriately formatted to ensure accurate parse trees are generated from the parser. In the case of the linguistic parser, the components of the NCPs are joined with the "underscore" to represent the phrase as one unit. Proper names are capitalized and front slashes (/) substituted with the word "or". Hyphenated and bracketed words are retained while double quotes are dropped and replaced by underscores to encase the terms. Acronyms are resolved into their full form.

For the concept-role mapping process, the fifty-five grammatical relations listed in Table 1 are considered inclusive of auxiliaries, arguments, modifiers and complements, extracted by the NLP parser. These grammatical relations capture semantic associations among adjacent and non-adjacent query constituents and are used to determine role-types of the query terms. Each query concept pair in the parse output, *RCP,* is tagged based on the set of predefined role type assignment that considers its functionality and detailed in Section 4.1. All concepts of a query are examined to determine whether there exist untagged or ambiguous concept role mappings. As detailed in Section 3.1.2, untagged query concepts occur when the most generic grammatical relation (the *dependent* relation) is formed due to an inability to define a more precise relation within the relational hierarchy. A statistical approach capitalizing on the frequency of a concept is used to determine untagged concept role types. The *Frequency Dependence* Case in handling untagged concepts in the *ConceptTag* structure is based on the two rules provided in Section 3.1.2: (i) the concept that occurs more frequently is the *CoI* and the other is a *DC*; (ii) if both concepts are equally frequent, both concepts are tagged as *CoIs*. Lastly, an *Inheritance Dependence* rule is used based on the fact that if a concept is tagged in one of the defined relations, the untagged concept inherits the concept role. As also described in Section 3.1.2, ambiguous concept role mapping occurs due to the nature of natural language queries in which adjacent and non-adjacent concepts may be syntactically or semantically associated. A *Ranked Relations* rule is defined to handle ambiguous concepts in the *ConceptTag* structure where the tag of the more significant role is retained according to the rules presented in Section 3.1.2.

### 4.2 Expansion Concept Extraction:
The lexical-semantic term pooling process is summarized in the algorithm below:

```
Input: Q ← The set of Web queries
1: for all q in Q do
2:    for all query terms in query do
3:       Determine correct sense of term through the disambiguation process
4:       Select base terms that are annotated as either CoIs or DCs
5:    end for
6: for all significant base terms (p) in query (q) do
7:    for each selected lexical-semantic relation(r)
8:       if r is synonymy
9:          Extract all synonyms
10:      if r is hypernymy
11:         Extract all hypernyms and their corresponding synonyms
12:      if r is hyponymy
13:         Extract all hyponyms and their corresponding synonyms
14:      if r is related-To
15:         Extract all related-To terms and their corresponding synonyms
16:   end for
17:   for each selected lexical-semantic relation pool ($p_r$)
18:      Identify multiple occurrences of identical terms, retain a single entry
19:      Compute semantic distance between each candidate term and all initial query terms
20:      Sort all candidate terms in descending order of semantic relatedness
21:   end for
22: end for
```

Sense disambiguation is performed to determine the correct meaning of a given term within the context of a query. For this purpose, a knowledge-based WSD technique is employed where overlaps within the definitions of two query terms are quantified via the Adapted Lesk semantic relatedness measure (Banerjee 2002). The set of base terms which will be used in the expansion term extraction process is then established based on the role-type of the query terms. Among all role-types, only terms which are annotated as *CoIs* and *DCs* are tagged as base terms. Candidate expansion terms are extracted into separate global pools consisting of extracted related terms from the four lexical-semantic term relationships (i.e. synonymy, hypernymy, hyponymy and *related-to*). All levels of the WordNet ontology are traversed to extract all relevant terms for each base term of a given query. The corresponding synonyms of each hypernym, hyponym and coordinate term are also included in the four global pools created. Each pool is then filtered to remove redundant candidate expansion terms generated for a single query. The semantic relatedness is computed to assess the global relatedness of a candidate expansion term to all original query terms and final pool of candidate terms are sorted in descending order.

### 4.3 Concept Weighting
The algorithm below describes the implementation of the concept weighting module:

```
Input: Q ← The set of Web queries
1: for all q in Q do
2:    InitializationOfGAParameters();
3:    String [ ] roleWeights = getGeneratedWeights()
4:    while (iterationNum != maxNumIterations) do
5:       Build input file for retrieval with generated weights
6:       tempMAP = getMAPfromToolkitOutputFile();
7:       if (tempMAP > maxMAP)
8:          Replace maxAP value with new MAP value
9: end for
```

Role type scores are predicted and optimized through the use of a GA and begins with the *InitializationOfGAParameters* method. The GA implementation for maximization of MAP includes the mutation and crossover genetic operators which are set at a rate of 10 and 1000 respectively. Role-type weights are generated for 4 out of 5 role types identified (i.e. *CoI*, *DC*, *RC* and *EC*) with a boundary condition set to restrict weights within the [0,1] interval. The *SC* role type is enforced with a weight of 0 to eliminate emphasis on general concepts as described earlier. A population of 200 chromosomes representing the role type weights is evolved over 100 iterations and serves as a termination condition. The mutation and crossover reproduction genetic operators are empirically set at a rate of 10 and 1000 respectively for maximum performance. The set of *roleWeights* generated are used in building the required input for the retrieval process via the Lemur Toolkit. Upon completion of the retrieval, *tempMAP* is assigned the MAP value of the query obtained from the toolkit output file. If at any iteration, the *tempMAP* value exceeds that of the *maxMAP* established in earlier iterations, the *maxMAP* value is updated with the new higher *MAP* value. The fitness value is used as a basis in the fitness proportionate selection method where a roulette-wheel selection technique is employed for choosing the chromosomes with high fitness value for reproduction and propagation to subsequent evolutions. The fitness value of each chromosome is boosted based on the overall MAP generated with the highest boost given to the highest MAP (above 0.5) across all iterations. The fitness value is also used to determine the appropriate number of iterations as the termination condition to limit the execution time. Upon determining the termination condition, the GA runs are executed for all queries.

## 5. EVALUATION
In this section, we introduce the test collections, detail the language-based processing experimental setup, introduce the compared frameworks then present the retrieval results. We conclude the section with a discussion of the obtained results.

### 5.1 Test Collections
The framework is tested on the Text REtrieval Conference (TREC) ad hoc test collections: Tipster Disk 1&2 consisting of documents from the Wall Street Journal (WSJ), Associated Press (AP), Federal Register (FR), Department of Energy (DOE) and Computer Select as well as GOV2 (.gov sites). These test collections were created as part of a text research project evaluation test bed by the National Institute of Standards and Technology (NIST). The collections used in this experiment range in size between 74, 520 documents to 25,114,919 documents (Table 5). For the purpose of evaluation, the datasets were sorted according to size and grouped as small (up to 100, 000 documents), medium (above 100, 000 documents and below 250,000 documents) and large (above 250, 000 documents). This organization of dataset sizes was determined based on the average size of Tipster datasets (approximately 205,785 documents) and the largest dataset (i.e. GOV2) used in this publication. The TREC ad hoc test topics consist of title and description fields which represent information needs (cf. Figure 4). In this study, the *title* field of the TREC 1, 3, 8 and Terabyte 2005 evaluation suites of the ad hoc retrieval test topics are examined. These query sets were chosen for their varying lengths and applicability to differing collection sizes. Each of these TREC evaluation suites consists of 50 topics (cf. Table 5).

Each topic has an associated set of binary judgments to indicate whether a document is relevant or irrelevant. These relevance judgments are used to assess the performance of the framework based on the number of relevant documents which are successfully retrieved.

**Table 5: Summary of TREC Collections & Topics**

| Dataset | # Docs | Topics | TREC |
|---|---|---|---|
| WSJ90_92 | 74,520 | 51-100 | 1 |
| AP88-90 | 242,918 | 51-100 | 1 |
| SJM1991 | 90,257 | 51-100 | 1 |
| WSJ87_92 | 173,252 | 151-200 | 3 |
| AP88_89 | 164,597 | 151-200 | 3 |
| Disk4&5 | 489,164 | 401-450 | 8 |
| GOV2 | 25,114,919 | 751-800 | Terabyte2005 |

```
<top>
<num> Number:  151
<title> Topic:  Coping with overcrowded prisons
<desc> Description:
```
The document will provide information on jail and prison overcrowding and how inmates are forced to cope with those conditions; or it will reveal plans to relieve the overcrowded condition.

`<narr>` Narrative:

A relevant document will describe scenes of overcrowding that have become all too common in jails and prisons around the country. The document will identify how inmates are forced to cope with those overcrowded conditions, and/or what the Correctional System is doing, or planning to do, to alleviate the crowded condition.

`</top>`

**Figure 4: Sample TREC Topic "*coping with overcrowded prisons*"**

### 5.2 Extrinsic Language-based Processing Setup
Grammatical relations are identified between query terms using the Stanford Parser (v1.6.8) and the role-types assigned as explained in Section 3.1. Incorporating lexical-semantic relations within the final expanded query requires several steps as defined in Section 3.2. Query terms that should be expanded are chosen on the basis of their functionality where terms annotated with the role-types *CoI* or *DC* were selected as base terms to be expanded. The WordNet ontology is adopted as the knowledge source for external knowledge-based processing as it provides an organized database of lexical-semantic relations and it is also recognized as the preferred knowledge source for the WSD process (Zhang & Gentile, 2011). WordNet consists of groups of terms arranged within synsets (bound by the lexical relation, synonym). Each synset is then linked to the corresponding semantically related synsets forming four separate hierarchies for each open-class part of speech noun, verb, adjective and adverb.

Disambiguation of polysemous terms was handled by the *WordNet::SenseRelate::AllWords (SR-AW)* tool. SR-AW determines the accurate sense of a given base term by considering a measure of relatedness between terms within a predetermined context window size. (Pedersen & Kolhatkar, 2009) showed that the performance of the Adapted Lesk (Banerjee 2002) was superior to other relatedness measures based on an ontological hierarchy. The Adapted Lesk method was also selected as it crosses parts-of-speech boundaries in the computation of relatedness as differing parts-of-speech are common among query terms. Therefore, this semantic relatedness measure was used for disambiguation within a default context window of size 3. The tool accepts three types of input: Raw, Tagged (Penn Treebank tags which are automatically reduced to the four basic parts-of-speech, i.e. noun, verb, adjective and adverb) and the wnTagged (only the four basic POS as assigned by WordNet). The set of queries previously tagged with their parts-of-speech by the Stanford NLP parser is fed to SR-AW via its Tagged input option to maintain consistency of the linguistic processes within the framework. Terms which are labelled with the suffixes #ND (i.e. term does not exist in WordNet), #NR (i.e. term is not related to surrounding terms), #o (i.e. closed-class terms), #IT (i.e. invalid tag), #MW (i.e. missing term) are ignored during the disambiguation process. The most significant impact to the disambiguation process is caused by the #ND and #NR suffixes which were predominantly found in the WSD results. Very often these suffixes are tagged to compound words, nouns and adjectives which are typically base terms used for expansion. A large number of such occurrences would cause the accuracy of the disambiguation process to be significantly reduced.

The #ND tag occurs under several conditions such as: (i) unrecognized NCPs and missing terms, (ii) inaccurate stemming and (iii) misclassified stop words. Unrecognized NCPs include not only missing NCPs due to the sparseness of NCPs within WordNet but also user-formed compound terms (e.g. US-USSR). It is beyond the scope of this paper to compensate for the sparseness of the ontology. The inaccurately stemmed terms issue was resolved with manual intervention to correct the root term to match the stem found within WordNet. Misclassified stop words are also sometimes tagged as #ND and this was handled by modifying the list of closed-class words recognized by the SR-AW to include all possible stop words. Since the number of remaining #ND tags after the issues were fixed was only about two to nine #ND tagged terms in each set of 50 queries, these terms were left unexpanded.

In the case of #NR tagged terms, terms could not be disambiguated due to a lack of relatedness to their surrounding terms. Varying the disambiguation window sizes showed that window size 2 resulted in a much larger number of #NR tags. There is no change in the number of #NR tags when utilizing windows sized 4 and 5. Here, the first sense of the term was assumed to be the correct sense. Through the extraction process defined in Sections 3.2 and 4.2, four global pools of lexical-semantic related terms are created for synonyms, hypernyms, hyponyms and coordinate terms. Breadth and depth of the WordNet ontology hierarchy are traversed to extract all relevant terms for each base term of a given query. The global relatedness of a candidate expansion term was then determined through a computation of average relatedness between each candidate expansion term and all base terms of a query in each pool using the Adapted Lesk semantic relatedness measure. This resulted in a ranked set of candidate expansion terms for each lexical-semantic related term pool. Morphological variants of terms were identified by stemming all candidate terms according to Porter's Stemming Algorithm (Porter, 1980).

The optimal number of expansion terms to be included is determined based on the findings of Greenberg (2001). An exploratory study to determine optimal thesauri-based query expansion processing methods in automatic and interactive environments was performed with real users and real queries. The author reports the number of expansion terms selected by users for each lexical-semantic relation, similar to those considered in this paper, ranges between 0.8 to 6.5 terms.

### 5.3 Retrieval Evaluation

To assess the performance of the proposed framework, it should experimentally be compared against recent research efforts in the field. However, the inconsistencies in the empirical setups of state-of-the-art frameworks prevent a fair comparison. It is acknowledged that the accepted de-facto standard for comparison are the unigram Language Model (LM) (Ponte & Croft, 1998) and Relevance Model (RM) (Lavrenko & Croft, 2001). An analysis of state-of-the-art works by Carpineto & Romano (2012) highlights large differences of baseline MAP results reported even on identical test sets. Accordingly, although the test collections are similar as far as queries processed and document set size are concerned, the produced baseline results are often different. The performance of the proposed technique cannot therefore be assessed by comparison against the absolute MAPs achieved in existing research. Therefore, in line with the practice of state-of-the-art research work, we examine the performance of our Language Resource-Based Query Expansion (LRBQE) framework by comparing against LM and RM. LRBQE is itself declined into four experimental variants: LRBQE_Syn, LRBQE_Hyper, LRBQE_Hypo and LRBQE_CT, corresponding to the coupled application of the intrinsic linguistic techniques with the spawning and integration of ontology-produced synonyms, hypernyms, hyponyms and coordinate terms respectively. The LM implementation produces the baseline results of the original queries without any modification. The RM performance itself translates the baseline performance of the state-of-the-art pseudo relevance feedback technique which considers the top n frequently occurring terms obtained from the top k documents returned for an initial query. A comparison is also made with a dependence-based query expansion (DQE) technique where base pairs are formed based on the notion of term dependencies (Huston & Croft, 2014), specifically sequential dependence where dependence is assumed to exist between adjacent query terms (Balaneshin-Kordan & Kotov 2017). The expanded query consists of original and expansion concepts weighted via GA.

As far as the evaluation metric is concerned, MAP is considered for analyzing the effectiveness of the compared frameworks similar to current research. MAP is the mean of the average precision scores for each query within the examined query pool. The paired t-test with 95% confidence level ($p < 0.05$) is performed to determine the statistical significance of differences in performance of the compared frameworks.

## 5.4 Results

*5.4.1 Effect of Using Synonyms as Expansion Terms*

The results of incorporating synonyms in the final expanded query are shown in Table 6. Statistically significant improvements in MAP over the baseline LM, RM and DQE are indicated by the symbols α, β and γ, respectively. Relative improvements in MAP are indicated in Figure 5.

**Table 6: Performance of LRBQE_Syn in Retrieval Tasks**

| Query No | Dataset | LM | RM | DQE | LRBQE_Syn |
|---|---|---|---|---|---|
| 51-100 | WSJ90-92 | 0.1874 | 0.2011 | 0.2178 | 0.2299 $^{αβγ}$ |
| | AP88-90 | 0.1979 | 0.2482 | 0.2314 | 0.2558 $^{αβγ}$ |
| | SJM1991 | 0.1463 | 0.1658 | 0.1741 | 0.1915 $^{αβγ}$ |
| 151-200 | WSJ87-92 | 0.2352 | 0.2874 | 0.2873 | 0.3357 $^{αβγ}$ |
| | AP88-89 | 0.2575 | 0.3252 | 0.3063 | 0.3558 $^{αβγ}$ |
| 401-450 | Disk 4-5 | 0.1926 | 0.2149 | 0.2124 | 0.2281 $^{αβγ}$ |
| 751-800 | GOV2 | 0.2944 | 0.3063 | 0.3221 | 0.34 $^{αβγ}$ |

The inclusion of synonyms in the final expanded query results in increased retrieval performance across all datasets over baselines, LM and RM. The observable relative improvements range between 3.1 to 16.8% over LM whilst a larger range of improvements were observed between 15.5 to 43% over RM.

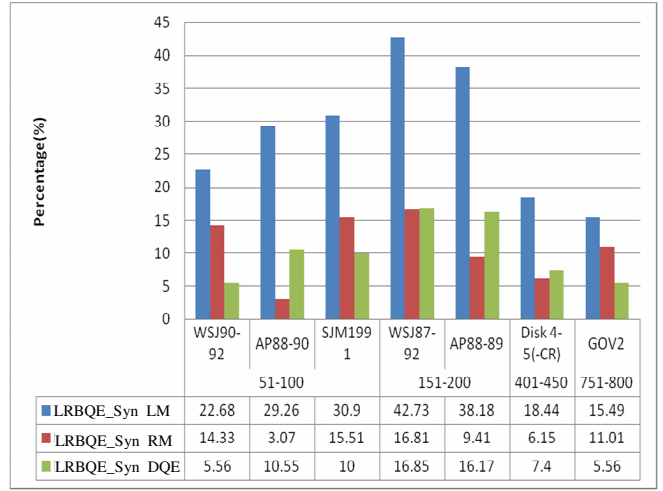

**Figure 5: Relative Performance of the LRBQE_Syn Variant**

Statistical significance is observed across all variations and datasets. These statistically significant improvements in retrieval performance are expected in line with the hypothesis that one of the leading causes of poor retrieval is the mismatch in the vocabulary chosen where varying terms are used to represent a single need. Also, expansion using synonyms results in relative improvements in MAP ranging from 5.56% to 16.85% across all datasets when compared with the performance of DQE. The improvements reported for all experimental variations are statistically significant.

*5.4.2 Effect of Using Hypernyms as Expansion Terms*

Retrieval performance achieved through the inclusion of hypernyms in the final expanded query is shown in Table 7.

**Table 7: Performance of LRBQE_Hyper in Retrieval Tasks**

| Query No | Dataset | LM | RM | DQE | LRBQE_Hyper |
|---|---|---|---|---|---|
| 51-100 | WSJ90-92 | 0.1874 | 0.2011 | 0.2178 | 0.2184 $^{αβ}$ |
| | AP88-90 | 0.1979 | 0.2482 | 0.2314 | 0.255 $^{αβγ}$ |
| | SJM1991 | 0.1463 | 0.1658 | 0.1741 | 0.1853 $^{αβγ}$ |
| 151-200 | WSJ87-92 | 0.2352 | 0.2874 | 0.2873 | 0.3404 $^{αβγ}$ |
| | AP88-89 | 0.2575 | 0.3252 | 0.3063 | 0.331 $^{αβγ}$ |
| 401-450 | Disk 4-5 | 0.1926 | 0.2149 | 0.2124 | 0.222 $^{αβγ}$ |
| 751-800 | GOV2 | 0.2944 | 0.3063 | 0.3221 | 0.3304 $^{αβγ}$ |

Statistically significant improvements in MAP over LM, RM and DQE are indicated by the symbols α, β and γ. The relative improvements in MAP are indicated in Figure 6. The expansion of queries using hypernyms resulted in increased retrieval performance across all datasets over baselines, LM and RM.

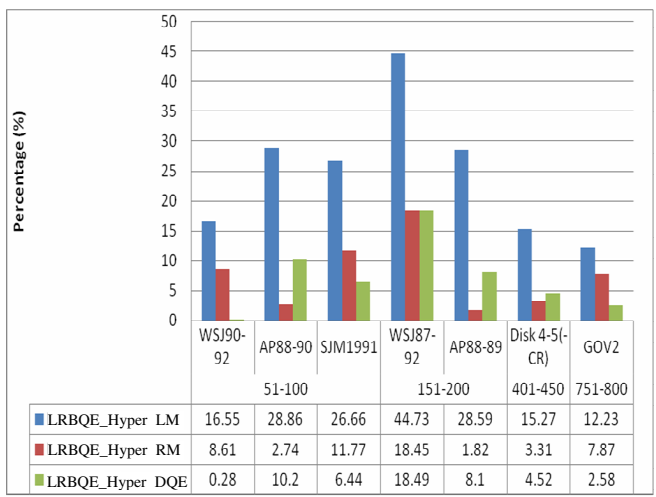

**Figure 6: Relative Performance of the LRBQE_Hyper Variant**

The increase in MAP ranges between 12.2 to 44.7% over LM whilst the range of improvements observed over RM was between 1.8 to 18.5%. These improvements in retrieval performance were all statistically significant across all variations and datasets. The LRBQE_Hyper variant also shows relative improvements across all experimental variations when compared to the DQE model where the improvements on six of the seven experimental variations are statistically significant. This reaffirms the method of hypernym selection used where the final expanded query is formulated with expansion terms that have the Top 5 highest global relation to all original query terms.

*5.4.3 Effect of Using Hyponyms as Expansion Terms*
Retrieval performance achieved through the inclusion of hyponyms in the final expanded query is shown in Table 8.

**Table 8: Performance of LRBQE_Hypo in Retrieval Tasks**

| Query No | Dataset | LM | RM | DQE | LRBQE_Hypo |
|---|---|---|---|---|---|
| 51-100 | WSJ90-92 | 0.1874 | 0.2011 | 0.2178 | 0.2164 $^{\alpha\beta}$ |
| | AP88-90 | 0.1979 | 0.2482 | 0.2314 | 0.2322 $^{\alpha}$ |
| | SJM1991 | 0.1463 | 0.1658 | 0.1741 | 0.1778 $^{\alpha\beta\gamma}$ |
| 151-200 | WSJ87-92 | 0.2352 | 0.2874 | 0.2873 | 0.328 $^{\alpha\beta\gamma}$ |
| | AP88-89 | 0.2575 | 0.3252 | 0.3063 | 0.3125 $^{\alpha\gamma}$ |
| 401-450 | Disk 4-5 | 0.1926 | 0.2149 | 0.2124 | 0.215 $^{\alpha\gamma}$ |
| 751-800 | GOV2 | 0.2944 | 0.3063 | 0.3221 | 0.3145 $^{\alpha\beta}$ |

Statistically significant improvements in MAP over LM, RM and DQE are indicated by the symbols α, β and γ. The relative improvements in MAP are indicated in Figure 7.

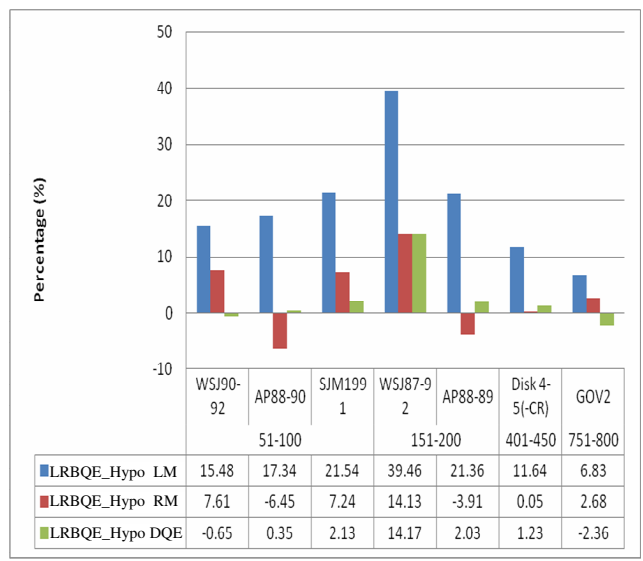

**Figure 7: Relative Performance of the LRBQE_Hypo Variant**

The use of hyponyms within the final expanded queries revealed a mixed outcome in terms of the achieved MAP. Overall, improvement in retrieval performance is observed over LM between 6.8 to 39.4% whilst changes in MAP over RM are both positive and negative. Deterioration in retrieval performance compared to RM is seen in two cases: datasets AP88-90 and AP88-89, whilst all other cases show improvement in retrieval performance averaging at 6.3%. When compared against DQE, relative improvements in retrieval performance average at 4% with four of the experiments being statistically significant. Also, deterioration in retrieval performance is observed on two datasets: WSJ90_92 and GOV2. The modest improvement in retrieval performance and in some cases deterioration is attributed to the fact that query terms may possess more than one hyponym. The problem with having multiple hyponyms is related to the meaning of hyponym type terms. Given that there may be several base terms used in the extraction process for a single query, there may also be multiple hyponyms for each query term. This issue cannot be easily resolved when dealing with automated query expansion as it is not possible to predict which of the many hyponyms may be of interest to a user. Improvement in retrieval performance may be observed in cases where the number of hyponyms related to a given base term is limited.

*5.4.4 Effect of Using Coordinate Terms as Expansion Terms*
Retrieval performance achieved through the inclusion of coordinate terms in the final expanded query is shown in Table 9.

Table 9: Performance of LRBQE_CT in Retrieval Tasks

| Query No | Dataset | LM | RM | DQE | LRBQE_CT |
|---|---|---|---|---|---|
| 51-100 | WSJ90-92 | 0.1874 | 0.2011 | 0.2178 | 0.2171 [αβ] |
| | AP88-90 | 0.1979 | 0.2482 | 0.2314 | 0.2346 [αγ] |
| | SJM1991 | 0.1463 | 0.1658 | 0.1741 | 0.1751 [αβ] |
| 151-200 | WSJ87-92 | 0.2352 | 0.2874 | 0.2873 | 0.3327 [αβγ] |
| | AP88-89 | 0.2575 | 0.3252 | 0.3063 | 0.3097 [αγ] |
| 401-450 | Disk 4-5 | 0.1926 | 0.2149 | 0.2124 | 0.2145 [α] |
| 751-800 | GOV2 | 0.2944 | 0.3063 | 0.3221 | 0.329 [αβγ] |

Statistically significant improvements in retrieval over LM, RM and DQE are indicated by the symbols α, β and γ, respectively and relative improvements in MAP are indicated in Figure 8. The inclusion of coordinate terms in the expansion process has resulted in some improvements as well as deterioration in retrieval performance across the datasets. Significant improvements are observed over LM within the range of 11.4 to 41.5%. Even though some improvements are observed on certain datasets, three tests showed deterioration over RM. When comparing with DQE, statistically significant improvements are only observed on four datasets with significant relative improvements averaging at 5%. Similarly to the impact of including hyponyms in the expanded queries, the use of coordinate terms results in the retrieval of irrelevant documents. The pool of terms considered in this form of expansion allows representing different aspects of a particular topic. This proves to be a difficult problem to resolve due to the inability to determine a user's possible preference in reformulation for an automated query expansion process. It may be possible to infer possible directions of reformulation by coupling the expansion process with personalization techniques examining users' search behaviour.

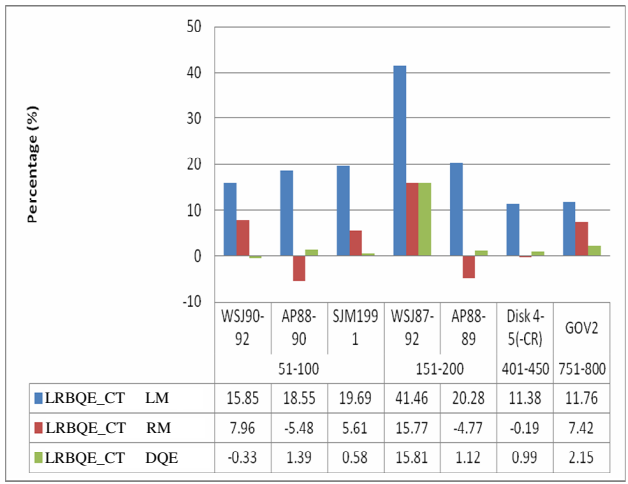

Figure 8: Relative Performance of the LRBQE_CT Variant

## 5.5 Discussion

Figure 9, 10 and 11 shows the performance improvement of the 4 lexical-semantic QE variations (i.e. synonym, hypernym, hyponym and coordinate terms) over baseline systems LM, RM and DQE.

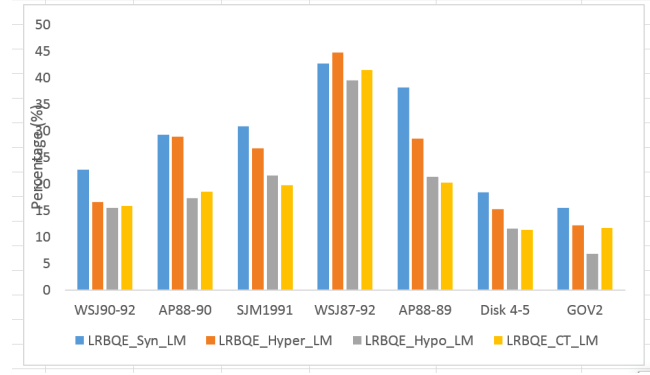

Figure 9: Performance of Lexical-Semantic QE Variations over LM

It is observed that the addition of lexical-semantic relations independently has shown significant improvements over the LM and RM frameworks especially for expansion with synonyms and hypernyms, answering one of the research issues on the usefulness of incorporating indirectly related terms in an expanded query. The improvements in MAP observed also validate the issue related to the use of lexical ontologies as a source of expansion terms. The LRBQE_Syn and LRBQE_Hyper empirical variants returned the largest range of increase over the baselines with most improvements being statistically significant. In the case of hyponyms (LRBQE_Hypo) and coordinate terms (LRBQE_CT), both increase and decrease in MAP were observed on specific datasets with the latter displaying more performance deterioration.

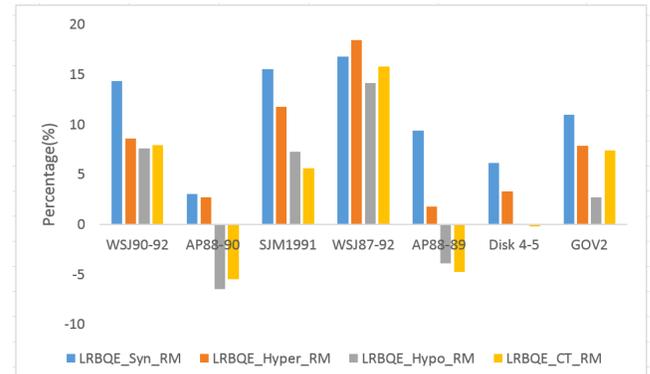

Figure 10: Performance of Lexical-Semantic QE Variations over RM

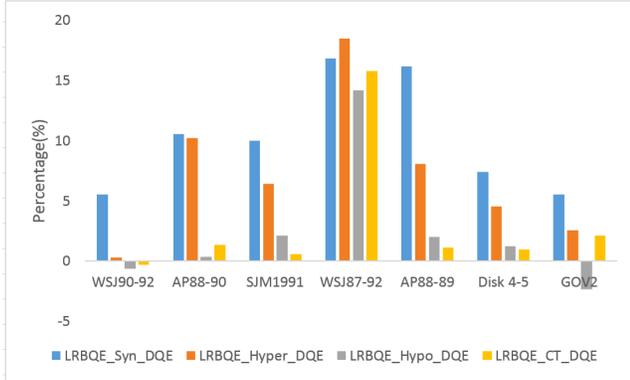

**Figure 11: Performance of Lexical-Semantic QE Variations over DQE**

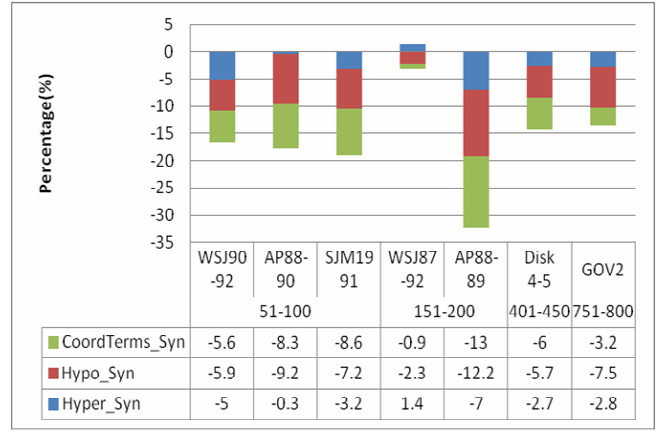

**Figure 12: Relative Performance of Language Resource-Based Expansion compared to LRBQE_Syn.**

Since expansion using synonyms (LRBQE_Syn) is the most generic form of lexical-semantic relation based expansion where the search goal is not altered by the process of expansion, the highest MAP is obtained in comparison to the expansion with hypernyms, hyponyms and coordinate terms. The three other variations deteriorate compared to LRBQE_Syn (cf. Figure 12). The LRBQE_Hyper variant shows an insignificant 1.4% increase on the WSJ87_92 dataset and a range of 0.3% to 7% decrease in MAP; LRBQE_Hypo shows negative change in MAP ranging between -2.3% to -12.2%; LRBQE_CT shows 0.9% to 13% drop in MAP, across all datasets.

Among the four best performing research frameworks highlighted by Carpineto & Romano (2012), only Liu et al. (2004) is an extrinsic knowledge-based processing technique. Despite having used the same Disk4&5 data collection as in our experiments, the exact query set examined was not clearly specified thus causing it to be not directly comparable. In the work of Cao et al. (2005) previously discussed, a link model is proposed that considers three semantic relations, namely synonymy, hypernymy and hyponymy. The equivalent dataset found in Cao et al. (2005) is the examination of topics 51-100 on the WSJ90_92, AP88_90 and SJM1991 datasets. In their work, relative improvements in MAP over the baseline LM model are reported between 5 to 11%. In the work of Dipasree et al. (2014) relative improvements of about 16% is seen over baseline RM. Comparatively, the four LRBQE methods proposed in this paper show a larger increment in retrieval performance ranging between 17 to 30%. The large difference in retrieval performance is attributed to the fact that in the work of Cao et al. (2005), the link model combines the contradicting hypernymy and hyponymy relations.

Regarding the drawbacks of using ontologies for query expansion, a noticeable limitation of WordNet is the limited coverage of concepts and phrases within the ontology. Query expansion via other external sources such as Wikipedia and search logs is of particular interest to fill the gaps of the WordNet ontology, if not replace it. This paper hypothesizes that extrinsic language resource based query expansion would benefit most in the case where multiple sources are utilized in the concept generation process. This is because ontologies such as WordNet provide semantically related expansion concepts whilst other corpus-based external sources rely on concept proximity to determine concept relatedness. This ensures that both direct and indirect links to a document are captured in the process of query expansion.

## 6. CONCLUSION

In this paper, we proposed a query expansion framework making use of intrinsic linguistic analysis to highlight the role of query concepts. Additionally, the use of an extrinsic knowledge-based resource made it possible to spawn concepts semantically coherent with the query content. In line with our postulation that grammatical pairing of query concepts, assessing the semantic relatedness of expansion concepts to the original query constituents and role-type based weighting are essential in query expansion, we have demonstrated improvements in retrieval performance with our proposed approach. Even though the expansion term spawning strategy can produce "noisy" concepts, experimentally the gain in terms of mean average precision with respect to the compared techniques is significant, especially in the variants considering the inclusion of synonyms and hypernyms. In our future work, we will further investigate gains in retrieval performance by studying cognitive patterns for expansion and eventually coupling statistical and extrinsic language-based resources to ensure only closely related concepts are included in the enriched query. Also, an application in multimedia web retrieval (Fauzi & Belkhatir, 2014) will be proposed enriching the works of Belkhatir (2011) and Fauzi & Belkhatir (2013).

# 7. REFERENCES

Alejandra Segura, N., Salvador-Sanchez, Garcia-Barriocanal, E., Prieto, M. (2011). An empirical analysis of ontology-based query expansion for learning resource searches using MERLOT and the Gene ontology. *Knowledge Base Systems*, 24(1), pp. 119-133

Balaneshin-Kordan, S., Kotov, A. (2017). Embedding-based query expansion for weighted sequential dependence retrieval model. *Proc. of ACM SIGIR*, pp. 1213-1216.

Banerjee, S. (2002). An adapted Lesk algorithm for word sense disambiguation using WordNet. *Computational Linguistics and Intelligent Text*, 136-145.

Belkhatir, M. (2011). A three-level architecture for bridging the image semantic gap. Multimedia Systems, 17(2), pp. 135-148

Bendersky, M., Croft, W. B. (2008). Discovering key concepts in verbose queries. *Proc. of ACM SIGIR*, pp. 491-498.

Bhogal, J., MacFarlane, A., Smith, P. (2007). A review of ontology based query expansion. *Information Processing & Management*, 43(4), pp. 866-886.

Bhogal, J., MacFarlane, A. (2013). Ontology based query expansion with a probabilistic retrieval model. *Proc. of 6th Information Retrieval Facility Conference (IRFC 2013)*

Cao, G., et al. (2005). Integrating word relationships into language models. *Proc. of ACM SIGIR*, pp. 298-305

Carpineto, C., Romano, G. (2012). A survey of automatic query expansion in information retrieval. *ACM Computing Surveys*, 44(1), pp. 1-50.

Covington, M.A. (2001). A fundamental algorithm for dependency parsing. *Proc. of the Annual ACM Southeast Conf.*, pp. 95-102

Cuadros, M., Rigau, G. (2006). Quality assessment of large scale knowledge resources. *Proc. of the Conf. on Empirical Methods in Natural Language Processing*, pp. 534–541.

Dipasree, P., Mitra, M., Datta, K. (2014), Improving query expansion using WordNet. *Journal of the Association for Information Science and Technology*, 65(12), pp. 2469-2478

Fauzi, F., Belkhatir, M. (2014). Image understanding and the web: a state-of-the-art review. Journal of Intelligent Information Systems, 43(2), pp. 271-306

Fauzi, F., Belkhatir, M. (2013). Multifaceted conceptual image indexing on the world wide web, Information Processing and Management, 49(2), pp. 420-440

Fogarolli, A. (2011). Wikipedia as a source of ontological knowledge: state of the art and application. *Intelligent Networking, Collaborative Systems and Applications*, 329, pp. 1-26.

Frank, E., et al. (1999). Domain-specific keyphrase extraction. *Proc. of IJCAI*, pp. 668-673

Greenberg, J. (2001). Optimal query expansion processing methods with semantically encoded structured thesauri terminology. *Journal of the American Society for Information Science*, 52, pp. 487-498.

Hollink, L., Schreiber, G., Wielinga, B. (2007). Patterns of semantic relations to improve image content search. Journal of Web Semantics 5, 3, 195-203

Huston, S., Croft, B.W. (2014). A comparison of retrieval models using term dependencies. *Proc. ACM CIKM*, pp. 111-120

Kim, S.-B., Seo, H.-C. Rim, H.-C. (2004). Information retrieval using word senses: root sense tagging approach. *Proc. of ACM SIGIR*, pp. 258 – 265.

Kraft D.H., Petry F.E., Buckles B.P., Sadasivan T. (1995). Applying genetic algorithms to information retrieval systems via relevance feedback. *In Fuzziness in Database Management Systems*. pp. 330-344.

Kuroda, K., Bond, F. (2010). Why Wikipedia needs to make friends with WordNet. *Proc. of the 5th International Conference on the Global Wordnet Association)*, pp.9-16.

Lavrenko, V., Croft, B.W. (2001). Relevance-based language models. *Proc. of ACM SIGIR*, pp. 120-127.

Lioma, C., Ounis, I. (2008). A syntactically-based query reformulation technique for information retrieval. *Information Processing & Management*, 44(1), pp. 143-162

Liu, S., Liu, F., Yu, C., Morgan, S. (2004). An effective approach to document retrieval via utilizing WordNet and recognizing phrases. *Proc. of ACM SIGIR*, pp. 266-272.

Liu, S., Yu, C., Meng, W. (2005). Word sense disambiguation in queries. *Proc. of ACM CIKM,* pp. 525–532

Maree, M., Belkhatir, M. (2011). A coupled Statistical/Semantic framework for merging heterogeneous domain-specific ontologies. *Proc. of Int. Conf. on Tools with Artificial Intelligence*, pp. 159-166

Maree, M., Belkhatir, M. (2013). Coupling semantic and statistical techniques for dynamically enriching web ontologies. *Journal of Intelligent Information Systems*, 40(3), pp. 455-478

Maree, M., Belkhatir, M. (2015). Addressing semantic heterogeneity through multiple knowledge base assisted merging of domain-specific ontologies. *Knowledge Base Systems*, 73, pp. 199-211

Marneffe, M.C., Manning, C.D. (2008). "Stanford typed dependencies manual." Stanford University, Technical report.

McCarthy, D., Carroll, J. (2003). Disambiguating nouns, verbs, and adjectives using automatically acquired selectional preferences. *Computational Linguistics*, 29(4), pp. 639-654.

Mihalcea, R. (2007). Using Wikipedia for automatic word sense disambiguation. *Proc. of HLT-NAACL*, pp. 196-203

Navigli, R., Velardi, P. (2002). An analysis of ontology-based query expansion strategies. *Proc. of the International Workshop on Adaptive Text Extraction and Mining*, pp. 42-49.

Paik, J.H., Oard, D.W. (2014). A fixed-point method for weighting terms in verbose informational queries. *Proc. of CIKM*, pp. 131-140

Park, J.H., Croft, B.W., Smith, D.A. (2011). A quasi-synchronous dependence model for information retrieval. *Proc. of CIKM*, pp. 17-26.

Pinter, Y., Reichart, R., Szpektor, I. (2016). Syntactic parsing of web queries with question intent. *Proc. of HLT-NAACL*, pp. 670-680.

Pedersen, T., Kolhatkar, V. (2009). WordNet :: SenseRelate :: AllWords - A broad coverage word sense tagger that maximizes semantic relatedness. *Proc. of Annual Conference of the North American Chapter of ACL*, pp. 17-20.

Pedersen, T., Patwardhan, S., Michelizzi, J. (2004). WordNet:: Similarity: measuring the relatedness of concepts. *Proc. of HLT-NAACL,* pp. 38–41.

Ponte, J. M., Croft, B.W. (1998). A language modeling approach to information retrieval. *Proc. of ACM SIGIR*, pp. 275–281.

Radhouani, S., Lim, J.H., Chevallet, J. P., Falquet, G. (2006). Combining textual and visual ontologies to solve medical multimodal queries. *Proc. of IEEE ICME,* pp. 1853–1856

Selvaretnam, B., Belkhatir, M. (2012). Human language technology and query expansion: issues, state-of-the-art and perspectives. *Journal of Intelligent Information Systems*, 38(3), pp. 709-740

Selvaretnam, B., Belkhatir, M., Messom, C. (2013). A coupled linguistics/statistical technique for query structure classification and its application to Query Expansion. *Proc. of FSKD*, pp. 1105-1109

Selvaretnam, B., Belkhatir, M. (2016). A linguistically driven framework for query expansion via grammatical constituent highlighting and role-based concept weighting. *Information Processing & Management*, 52(2), pp. 174-192

Simon, P., Sathya, S. (2009). Genetic algorithm for information retrieval. *Proc. of International Conference on Intelligent Agent & Multi-Agent Systems*

Song, R., et al. (2008). Viewing term proximity from a different perspective. *Proc. of ECIR*, pp. 346-357

Tudhope, D., Alani, H., Jones, C. (2001). Augmenting thesaurus relationships: Possibilities for retrieval. *Journal of Digital Information* 1, 8

Tuominen, J., Kauppinen, T., Viljanen, K., Hyvönen, E. (2009). Ontology-based query expansion widget for information retrieval. *Proc. of 5th Workshop on Scripting and Development for the Semantic Web*

Voorhees, E.M. (1994). Query expansion using lexical-semantic relations. *Proc. of ACM SIGIR*, pp. 61-69

Yang, J., Korfhage, R., Rasmussen, E. (1992). Query improvement in information retrieval using genetic algorithms--a report on the experiments of the TREC project. *Proc. of TREC-1*, pp. 31-58.

Zhang, Z., Gentile, A.L., Ciravegna, F. (2011). Harnessing different knowledge sources to measure semantic relatedness under a uniform model. *Proc. of EMNLP*, pp. 991-1002.

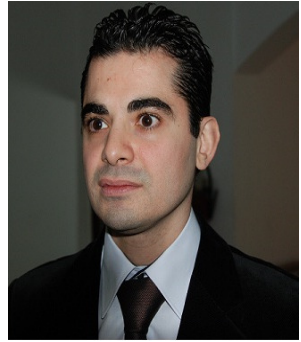

Mohammed Belkhatir

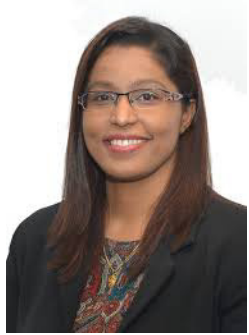

Bhawani Selvaretnam